\documentclass[letterpaper, 10 pt, conference]{ieeeconf}  % Comment this line out if you need a4paper

\IEEEoverridecommandlockouts                              % This command is only needed if 
\DeclareUnicodeCharacter{0301}{*************************************}% Needed to meet printer requirements. 

\usepackage{amsmath} % assumes amsmath package installed
\usepackage{xcolor}
\usepackage{graphicx}
\usepackage{tabularx}
\usepackage{booktabs} %%,multirow,array}
\usepackage{multirow}
\usepackage{graphicx} 
\usepackage{caption,subcaption}
\usepackage{todonotes}

\title{\LARGE \bf
Frontier Shepherding: A Bio-inspired Multi-robot Framework for Large-Scale Exploration
}

\author{John Lewis$^{1}$, Meysam Basiri$^{1}$, and Pedro U. Lima$^{1}$% <-this % stops a space
\thanks{* This work was supported by doctoral grant from Fundação para a Ciência e a Tecnologia (FCT) UI/BD/153758/2022, Aero.Next project (PRR - C645727867-
00000066) and ISR/LARSyS Strategic Funding through the FCT project
DOI: 10.54499/UIDB/50009/2020, DOI: 10.54499/UIDP/50009/2020, DOI: 10.54499/LA/P/0083/2020}% <-this % stops a space
\thanks{$^{1}$J. Lewis, M. Basiri and P. Lima are with the Institute for Systems and Robotics,
Instituto Superior T\'ecnico, Universidade de Lisboa, Lisbon 1049-001, Portugal
(e-mail: \{\textit{john.lewis; meysam.basiri; pedro.lima}\}@\textit{tecnico.ulisboa.pt}).}%
}
\usepackage[
backend=biber,
style=numeric-comp,
sorting=none
]{biblatex}
\bibliography{IEEEexample}

\newcommand\blfootnote[1]{%
  \begingroup
  \renewcommand\thefootnote{}\footnote{#1}%
  \addtocounter{footnote}{-1}%
  \endgroup
}

\usepackage{makecell}
\begin{document}

\maketitle
\thispagestyle{empty}
\pagestyle{empty}

%%%%%%%%%%%%%%%%%%%%%%%%%%%%%%%%%%%%%%%%%%%%%%%%%%%%%%%%%%%%%%%%%%%%%%%%%%%%%%%%
\begin{abstract}

%Gig
%Efficient exploration of large-scale environments remains a critical challenge in robotics, with applications ranging from environmental monitoring to search and rescue operations. This article proposes a bio-intelligent multi-robot framework, \textit{Frontier Shepherding (FroShe)}, for large-scale exploration. The presented bio-inspired framework heuristically models frontier exploration similar to the shepherding behavior of herding dogs. This is achieved by modeling frontiers as a sheep swarm reacting to robots modeled as shepherding dogs. The framework is robust across varying environment sizes and obstacle densities and can be deployed across multiple agents with minimal parameter tuning. The simulation results demonstrate that the proposed method performed consistently, regardless of the varying sizes and obstacle densities of the simulated environment. With an increase in the number of agents, the proposed method outperforms other state-of-the-art exploration methods, with an average improvement of $20\%$ with the next-best approach (for $3$ UAVs). The proposed technique was implemented and tested in a single- and dual-drone scenario in a forest-like real-world environment.
Efficient exploration of large-scale environments remains a critical challenge in robotics, with applications ranging from environmental monitoring to search and rescue operations. This article proposes Frontier Shepherding (FroShe), a bio-inspired multi-robot framework for large-scale exploration. The framework heuristically models frontier exploration based on the shepherding behavior of herding dogs, where frontiers are treated as a swarm of sheep reacting to robots modeled as shepherding dogs. FroShe is robust across varying environment sizes and obstacle densities, requiring minimal parameter tuning for deployment across multiple agents. Simulation results demonstrate that the proposed method performs consistently, regardless of environment complexity, and outperforms state-of-the-art exploration strategies by an average of $20\%$ with three UAVs. The approach was further validated in real-world experiments using single- and dual-drone deployments in a forest-like environment.

%In this article, a multi-robot large-scale exploration framework, \textit{Frontier Shepherding (FroShe)}, is proposed. The presented bio-inspired framework models frontier exploration similar to the shepherding behavior of herding dogs. This heuristic approach to frontier exploration is characterized by its robustness across varying environments and ease of deployment across multiple agents. Furthermore, frontier behavior is modeled as a sheep swarm to tackle possible communication and computational delays hindering exploration. Simulation results in environments showcase the proposed method consistently performed irrespective of varying sizes and occlusion. With the increase in the number of agents, the proposed method outperforms other state-of-the-art exploration methods, with an average improvement of $20\%$ with the following best approach(for $3$ UAVs). The proposed technique was implemented and tested in a single and dual drone scenario in a forest-like environment.

\end{abstract}

%%%%%%%%%%%%%%%%%%%%%%%%%%%%%%%%%%%%%%%%%%%%%%%%%%%%%%%%%%%%%%%%%%%%%%%%%%%%%%%%
\section{INTRODUCTION}

\label{sec:introduction}
%A truly autonomous
%Robotic applications involving exploration or surveillance of a large-scale area can be taxing for a single robot. Delegation of such tasks across multiple robots improves overall efficiency. However, multi-robot solutions often require prior planning, training, or optimization techniques to improve coordination, minimize overlapped exploration, and overcome communication constraints. Such planning might be complex in scenarios such as search and rescue, disaster response, and large-scale exploration of unknown, unstructured areas. Quick and robust deployment of multiple robots for such scenarios while ensuring complete coverage is thus vital.

Robotic applications involving the exploration and surveillance of large-scale environments present significant challenges, particularly when relying on a single robot. Distributing such tasks across multiple robots enhances efficiency but often necessitates prior planning, training, or optimization to improve coordination, minimize redundant exploration, and address communication constraints. These challenges become even more pronounced in scenarios such as search and rescue, disaster response, large-scale mapping of unknown terrains, and planetary exploration, where rapid and robust deployment is critical. Ensuring complete and efficient coverage in such dynamic and unstructured environments requires adaptable multi-robot strategies that can operate with minimal prior information and computational overhead.

%This paper presents a novel multi-robot exploration framework that exploits the simplicity of bio-mimetic behaviors to explore unknown areas. The proposed modular, online, and decentralized strategy enables robust and scalable exploration suitable for quickly deploying computationally constrained robot(s) for mapping unknown and hazardous terrains. The outline of the paper is as follows: Section \ref{sec:literature_review} provides the literature review on various robotic exploration strategies and the use of bio-mimetics in robotics. The proposed multi-agent exploration framework is discussed in Section \ref{sec:methodology}. The experiments and results are presented in Section \ref{sec:results}. Finally, Section \ref{sec:conclusion} concludes the findings and pitches possible improvements to the proposed method. 

%This paper presents a novel multi-robot exploration framework that exploits the simplicity of bio-mimetic behaviors to explore unknown areas. 

\subsection{Related Work}

\label{sec:literature_review}
%Fast and robust autonomous exploration is an essential aspect of outdoor robotics, crucially to achieve complete autonomy in scenarios such as search and rescue \cite{basiri2021multipurpose}, disaster response \cite{ghassemi2022multi}, and mapping of large-scale unknown areas \cite{bartolomei2023fast}. A conventional autonomous exploration task is carried out by defining \textit{frontiers} \cite{bircher2016receding} as the boundary between the known/mapped and unknown/unmapped areas.  Exploration is achieved by pushing this boundary or mathematically by minimizing the perimeter or length of the frontier.  However, this straightforward solution becomes inadequate and increasingly complex in the presence of obstacles, energy and time constraints, and completeness of the exploration. Thus, prioritizing viewpoints from the knowledge of frontier, the path planned and the perception sensor can optimize the flight time required for exploration \cite{bartolomei2023fast,bircher2016receding,zhou2016fast}, where viewpoints are the points at which the frontier can be altered.

Fast and robust autonomous exploration is essential for outdoor robotics, particularly in applications such as search and rescue, \cite{basiri2021multipurpose}, disaster response \cite{ghassemi2022multi}, and mapping of large-scale unknown areas \cite{bartolomei2023fast}. A common approach to autonomous exploration involves defining \textit{frontiers}, as the boundary between known/mapped and unknown/unmapped areas, and systematically pushing this boundary outward to expand the known area. This can be formulated as minimizing the perimeter of the frontier or selecting viewpoints that optimize exploration efficiency. However, this straightforward approach becomes increasingly complex in the presence of obstacles, energy and time constraints, and the need for complete coverage. Prioritizing viewpoints based on frontier knowledge, planned paths, and sensor capabilities can significantly optimize the flight time required for exploration \cite{bartolomei2023fast,bircher2016receding,zhou2016fast}.

Frontier exploration strategies include greedy exploration \cite{duberg2022ufoexplorer}, proximal exploration \cite{cieslewski2017rapid} or exploration based on prior training \cite{hu2020voronoi}. Proximal exploration strategies require conditional supervision to prevent the robot from being stuck in local minima. In contrast, greedy exploration strategies can often lead to sub-par time management, especially in highly cluttered environments. Deep learning or reinforcement learning strategies require prior training and substantial data, resulting in slow deployment. Furthermore, reliance on training on extensive data may not capture the complexities of unforeseen environments. Fast deployment is crucial in disaster response and search and rescue scenarios. Markov decision process (MDP) based exploration strategies  \cite{budd2020markov,barbosa2021risk,budd2023bayesian} often guarantee safety but can be computationally intensive.

%An attractive solution to meeting time is to increase the speed of the exploring robot. However, this is a safety concern, especially in an unknown environment. 
Increasing the speed of exploration robots may offer benefits in terms of time efficiency but introduces several challenges. These include increased collision risks, limitations in perception and sensing, control and stability issues, and the need for faster data processing and communication. An alternative solution is to distribute the overall exploration task across multiple robots. Over the years, multi-agent exploration solutions have evolved from general strategies \cite{burgard2000collaborative,burgard2005coordinated} to more specific approaches for tasks such as forest \cite{bartolomei2023fast}, cave \cite{petravcek2021large} and  indoor \cite{lu2021multi} exploration. Bartolomei et al. \cite{bartolomei2023fast} enhance exploration by employing a dual-mode approach, consisting of the explorer and collector roles. In explorer mode, agents push the frontiers, while in collector mode, they focus on exploring the "trails" or leftover unknown pockets. This dual mode enables variable velocity, with higher speeds in collector mode to capture the "trails" more quickly, thereby ensuring both speed and safety.

%enhance exploration by utilizing a dual mode, namely explorer and collector. In explorer, the agents push the frontiers, and in collector, the "trails" or leftover unknown pockets are explored. The dual mode enables a variable velocity, with a higher velocity in collector mode to capture the "trails" faster, thereby guaranteeing speed and safety.
% \cite{bartolomei2023fast} further enhances the algorithm by utilizing a dual mode of exploration, namely explorer and collector. In explorer, the agents push the frontiers and in collector, the "trails" or left over unknown pockets are explored. The dual mode enables  \cite{bartolomei2023fast} to have a variable velocity with a higher velocity in collector mode to capture the "trails" faster, thereby guaranteeing speed and safety. 

%However, multi-robot exploration comes at the cost of coordination requisites and communication constraints.

%An uncoordinated multi-agent system can often lead to multiple sweeps of the same area to achieve complete coverage. Thus, communication and coordination are vital to extract the full effectiveness of a multi-robot system. This requirement increases complexity with the increase in number of robots. 
Communication and coordination are essential for maximizing the effectiveness of a multi-robot system. This requirement becomes more complex as the number of robots increases. 
In a communication-constrained scenario, Yuman Gao et al. \cite{gao2022meeting} proposed a framework where the agents initially coordinate a meeting point to share the map. The agents then explore an area and reconvene at the predetermined meeting point to merge maps and determine the meeting points and areas to explore. In bandwidth-constrained instances, sparse information can be transmitted over a long-range communication protocol, and complete information transfer can be instigated utilizing a short-range communication protocol. In such scenarios, the method proposed by Lewis et al. \cite{lewis2022collaborative} enables sharing minimal pointcloud, and GNSS coordinates over long-range communication, while short-range communication is used for sharing a complete map. This minimal information can assist in exploration as the robots are clued in on the possible explored areas. %\cite{bone2023decentralised} provides a coordinated exploration utilizing a decentralized approach by relying on point-to-point communication. Such a coordinated approach ensures that the agents are spread out, thereby minimizing the overlap with previously explored areas of other agents.
A coordinated exploration utilizing a decentralized approach by relying on point-to-point communication \cite{bone2023decentralised} ensures that the agents are spread out, thereby minimizing the overlap with previously explored areas of other agents.

Naturally occurring behaviors such as sheep herding \cite{strombom2014solving}, ant foraging \cite{dorigo2006ant}, and fish swarming \cite{couzin2002collective} can be heuristically modeled to incorporate swarm-like behavior. These minimal heuristic inter-agent relations lead to emergent behaviors and can be used for rapid deployment and control of robotic swarms \cite{spears2005overview}. Additional swarm control can be attained by integrating heuristics into the agent behavior, which acts as a reaction to an external agent. The added change in swarm dynamics can be predatory  \cite{strombom2014solving,benoit2009cooperative} or leader-like  \cite{dorigo2018influence,goel2019leader}, depending on the nature of the task at hand. The ability to control a large swarm of robots by manipulating a few robots is a favorable option as it minimizes the control, communication, and coordination requirements. In the context of exploration, prior works \cite{dirafzoon2013topological,goel2019leader,zhi2021learning} have explored or can be extended to include exploration tasks utilizing swarms.

%In this paper, we present Frontier Shepherding (\textit{FroShe}), a novel bio-mimetic multi-robot framework for large-scale exploration. The proposed exploration framework, \textit{(i)} utilizes heuristic bio-mimetism to explore frontiers; \textit{(ii)} achieves frontier prioritization using virtual bio-mimetic agents and behavior; \textit{(iii)} provides robust performances across varying environments and coverage areas.   %, mention the reliance and basic stuff...}

\subsection{Contributions} \label{sec:contributions} \blfootnote{Video: https://youtu.be/Kme95Bf8ros}
%In this article, we present Frontier Shepherding (\textit{FroShe}), a novel bio-inspired multi-robot framework for large-scale exploration. The proposed framework, \textit{(i)} utilizes heuristic bio-mimetism to explore frontiers; \textit{(ii)} achieves frontier prioritization using virtual bio-mimetic agents and behavior; \textit{(iii)} provides robust performances across varying environments and coverage areas. The proposed modular, online, and decentralized strategy enables adaptable, robust and scalable exploration suitable for quickly deploying computationally constrained robot(s) for mapping unknown and hazardous terrains.    %, mention the reliance and basic stuff...}
This paper introduces Frontier Shepherding (FroShe), a novel bio-inspired multi-robot exploration framework. The key contributions are:

\begin{itemize}
    \item \textbf{Bio-mimetic Frontier Exploration}: A heuristic approach that models frontiers as a dynamic swarm, enabling efficient multi-robot exploration inspired by natural shepherding behaviors.
    \item \textbf{Adaptive Frontier Prioritization}: Virtual bio-mimetic agents dynamically represent and prioritize frontiers, enhancing exploration efficiency.
    \item \textbf{Scalable and Decentralized Deployment}: A modular, online framework that enables robust multi-robot exploration across varying environments with minimal parameter tuning.
    \item \textbf{Comparative Evaluation in Simulations}: Extensive Software-In-The-Loop (SITL)  simulations show that FroShe achieves consistent performance across different scenarios and agent configurations.
    \item \textbf{Real-World Validation}: Implementation in a forest-like environment with single- and dual-UAV setups demonstrates the framework’s practicality and robustness.
    
\end{itemize}

\subsection{Outline}
The article's outline is as follows: Section \ref{sec:methodology} details the proposed multi-agent exploration framework. The experiments and results are presented in Section \ref{sec:results}. Finally, Section \ref{sec:conclusion} concludes the findings and pitches possible improvements to the proposed method.

%%%%%%%%%%%%%%%%%%%%%%%%%%%%%%%%%%%%%%%%%%%%%%%%%%%%%%%%%%%%%%%%%%%%%%%%%%%%%%%%

%%%%%%%%%%%%%%%%%%%%%%%%%%%%%%%%%%%%%%%%%%%%%%%%%%%%%%%%%%%%%%%%%%%%%%%%%%%%%%%%

\section{METHODOLOGY}

\label{sec:methodology}
%A team of $n_r$ robots, $\mathcal{R} =[R_1\ldots,R_{n_r}]$,  is tasked to explore and map an unknown environment of area, $\mathcal{A}$. The robots are equipped with a $3$D-sensor, with perception range $L$,  for generating a $3$D terrain representation. The proposed methodology is broadly grouped into three stages, as shown in Fig.  \ref{fig:Methodology}. %The robots are equipped with a $3$D-sensor for map generation as the agent traverses the terrain. %Each robot's map is shared with the rest of the team to generate a global view as perceived by the whole team.

\begin{figure}[h]
    \includegraphics[width=0.5\textwidth]{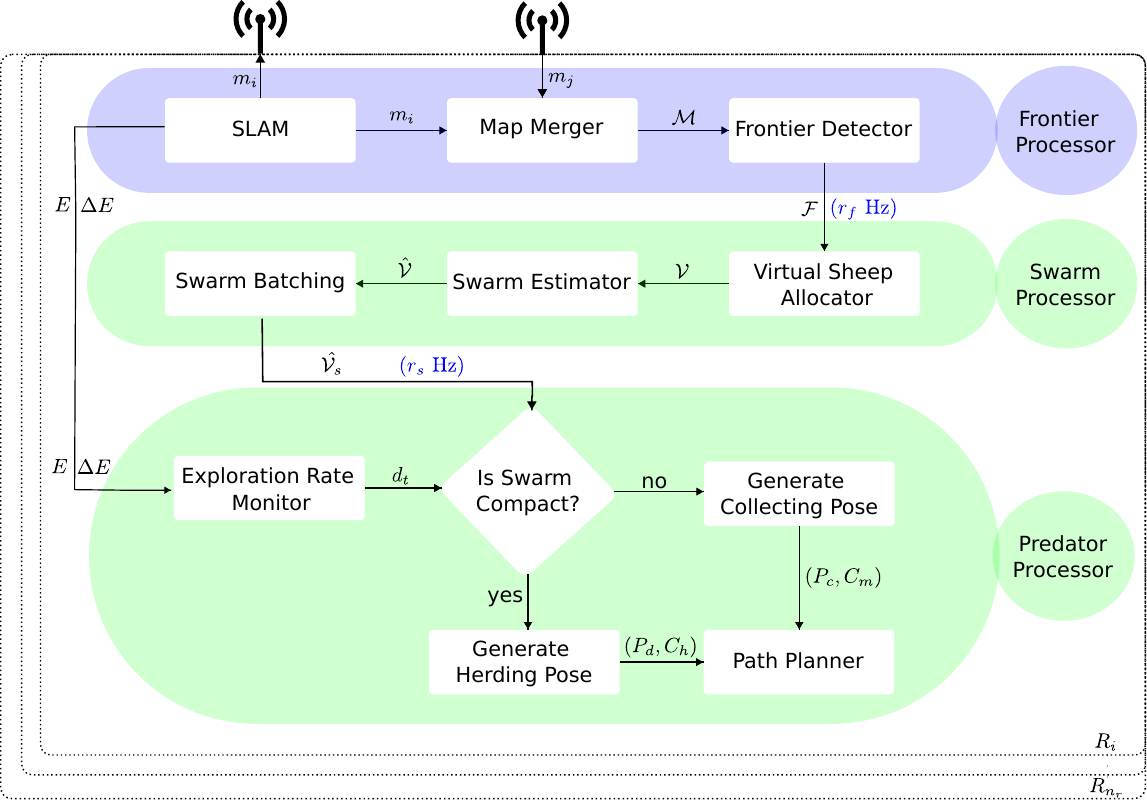}
    \captionsetup{justification=centering}
    \caption{The FroShe framework. The key contributions of the proposed method are highlighted in green.}
    \label{fig:Methodology}
\end{figure}

A team of $n_r$ robots, $\mathcal{R} =[R_1\ldots,R_{n_r}]$,  is tasked to explore and map an unknown environment of area, $\mathcal{A}$. Each robot is equipped with a perception sensor (e.g., Ouster, Intel RealSense), with a perception range of $L$, enabling it to generate a local representation of the terrain. The proposed methodology is broadly grouped into three stages, as shown in Fig.  \ref{fig:Methodology}.

% \begin{align}
% &&I_{i}(t+1) &= h\dot{\bar{S}}_i(t),&& \label{Inertia} \\
% &&F_{i} &= \dfrac{\rho_s}{||\bar{P_j}-\bar{S_i}||} \quad ,\forall \, P_i \in \mathcal{P},&&\label{Predator}\\
% &&C_{i} &= \dfrac{c}{n}\sum_{j=1}^{n} \bar{S}_j, &&\label{Cluster}\\
% &&Re_i &= \begin{cases}
%     \rho_a \dfrac{(\bar{S}_i-\bar{S}_j)}{||\bar{S}_i-\bar{S}_j||} &,\forall |\bar{S}_i-\bar{S}_j| \leq R_a,  \\
%     0&,\text{otherwise},
%   \end{cases}&&\label{Repulsion}\\
% &&\dot{\bar{S}}_{i}(t+1) &= E_i = p.e.v_s \quad \forall \, S_i \in \mathcal{S}_g,&&\label{error}\\
% &&\dot{\bar{S}}_{i}(t+1) &=\bar{S}_{i}(t+1)-\bar{S}_{i}(t),&&\label{delta}\\
% &&\dot{\bar{S}}_{i}(t+1) &= v_s(I_{i}+F_{i}+C_{i}+Re_i) \quad \forall  \, S_i \in \mathcal{S}_p.&&\label{Resultant}
% \end{align}

\subsection{Frontier Processor}\label{sec:frontierprocessing}
Each robot $R_i\in\mathcal{R}$, with pose $\bar{R_i}$, runs a continuous onboard mapping algorithm \cite{reinke2022locus,palieri2020locus,shan2020lio} to generate and update its own map%\cite{hornung2013octomap}
, $m_i$. In the absence of a global communication topology, a communication-constrained map merging algorithm \cite{lewis2022collaborative,gao2022meeting} is used to merge $m_i$ with $m_j$ (for all $j\neq i$), forming a shared global representation $\mathcal{M}$. The set of discrete frontiers $\mathcal{F} = [f_1,f_2,...,f_{n_f}]$, is then extracted along the known-unknown boundary of $\mathcal{M}$ \cite{keidar2014efficient,senarathne2013efficient}. 
$\mathcal{M}$, $n_f$ and $\mathcal{F}$ are continuously updated, at a rate of $r_f$ Hz, throughout the exploration process.
% \todo[inline]{Previously: The set of nf map
% frontiers, F = [f1, f2 , ..., fnf ], in M is calculated [32, 33].
% F and M are continuously updated, at a rate of rf Hz,
% throughout the exploration process.% Reviewer 1: It would be helpful to have a short description of frontiers, and why there are multiple, , Where is $n_f$ determined? By the number of robots?} 
The modular structure of this frontier processing stage allows flexibility in selecting specific methods for SLAM, map merging, and frontier detection, depending on the available communication bandwidth, robot capabilities, and cooperation level. Given the computationally intensive modules of SLAM, communication delays, map merging, and frontier detection, the output of the frontier processor is expected to be slow and encounter delays in reality.

%\subsubsection{SLAM and Frontier Processing}
%We acquire the map, $M$, and the localization estimate, $O$ in $M$, from the online SLAM module. A 2D occupancy grid, $O_c$, is generated from $M$. Using $O_c$ and $O$, we determine the frontiers, $F$, that need to be explored. Frontier acquisition can be further improved with the use of binary octomaps \cite{hornung13auro} and related algorithms \cite{zhu20153d,batinovic2021multi}. However, we are limiting our frontier search to a 2D occupancy grid in an effort to keep the overall complexity to a minimum \checkplease{and focus primarily on the frontier-processing presented in the following sections}. We leave enhancements in this module for future work.route

\subsection{Swarm Processor}\label{methodology:swarmprocessor}

%To model frontier exploration in a bio-inspired manner, we draw an analogy to shepherding behavior \cite{strombom2014solving}, where a team of robots (\(\mathcal{R}\)), explores frontiers (\(\mathcal{F}\)), similarly to how shepherding dogs (\(\mathcal{P}\)) herd a flock of sheep (\(\mathcal{S}\)).

The task of herding a flock of sheep (\(\mathcal{S}\)) with shepherding dogs (\(\mathcal{P}\))—as studied in \cite{strombom2014solving}—serves as inspiration for modeling the exploration of frontiers (\(\mathcal{F}\)) by a team of robots (\(\mathcal{R}\)). 
%Building on this concept, we adapt and extend the shepherding principles to suit multi-robot exploration, enabling robots to estimate frontier behavior under delayed frontier processor outputs while enhancing exploration efficiency through bio-inspired heuristics.
We extend these shepherding principles to multi-robot exploration, enabling robots to estimate frontier behavior despite delays in the frontier processor while enhancing exploration efficiency through bio-inspired heuristics.

In the real-shepherding model described in \cite{strombom2014solving}, the movement of sheep is governed by five forces: inertial, erroneous, inter-agent repulsion, clustering, and predatory-response forces. Of these, inertial, erroneous, and repulsion forces continuously influence each sheep, while clustering and predatory-response forces activate only when a sheep is within a predator’s sphere of influence. To adapt this model for frontier exploration, we define a virtual sheep swarm $\mathcal{V}$ obtained from the frontiers, and robots act as shepherds.
Table I summarizes the forces applied in the original model and their adaptation for frontier representation. $h$, $\rho_a$, $c$, and $\rho_s$ are constants that determine the strength of each force that acts upon sheep $S_i$, with pose $\bar{S_i}$ and shepherds' pose $\bar{P_j}$. $e$ is a small erroneous constant emulating noise in the swarm.

%the behavior of the sheep swarm is heuristically emergent from $5$ forces. Out of these $5$ forces, inertial, erroneous, and inter-agent repulsion forces consistently act upon each sheep, while clustering and predatory-response forces are triggered when a sheep is within the sphere of influence of a predator. The various forces involved in the heuristic model \cite{strombom2014solving} for a swarm of $n$ sheep ($\mathcal{S}= [S_1\ldots,S_{n}]$)  in the presence of $m$ shepherding dogs ($\mathcal{P}= [P_1\ldots,P_{m}]$)  is summarized in \textbf{sheep} column of Table \ref{tab:analogy_table}. $h$, $\rho_a$, $c$, and $\rho_s$ are constants that determine the strength of each force that acts upon sheep $S_i$, with pose $\bar{S_i}$ and shepherds' pose $\bar{P_j}$. $e$ is a small erroneous constant emulating noise in the swarm. 

\begin{table}[h]
    \centering
    \begin{tabular}{|c|c|c|}
    \hline
       \textbf{Force}  & \textbf{on sheep, $S_i$} \cite{strombom2014solving} & \textbf{on virtual sheep, $v_i$}\\ \hline 
       Inertial Force& $h\dot{\bar{S}}_i(t)$ & $0$ \\  \hline
       \makecell{Inter-Agent\\ Repulsive Force}   & $\rho_a \frac{(\bar{S}_i-\bar{S}_j)}{||\bar{S}_i-\bar{S}_j||}$ & $f_{res}$\\ \hline
       Erroneous  Force& $e$ & $e$  \\ \hline
       Clustering  Force& $\frac{c}{n}\sum_{j=1,\neq i}^{n} \bar{S}_j$ &  $\frac{-c_f}{n}\sum_{j=1,\neq i}^{n_v} {v}_j$  \\ \hline 
       Predatory Force  & $\sum_{j=1}^{m} \frac{\rho_s}{||\bar{P_j}-\bar{S_i}||}$    & $\sum_{j=1}^{n_r} \frac{\rho_f}{||\bar{R_j}-{v_i}||}$ \\ \hline 
       \makecell{Predator\\ detection range} & $r_p$ & $L$ \\ 
    \hline
    \end{tabular}
    \caption{Modelling frontiers analogous to sheep swarm. The cumulative sum of the forces determines the behavior of the $S_i$ and $v_i$.}
    \label{tab:analogy_table}
\end{table}

The swarm processor module heuristically models $\mathcal{F}$ as a virtual sheep swarm at a rate of $r_s$ where $(r_s >>r_f)$ Hz. In the following submodules, the parameters of this heuristic model are tuned to replicate frontier behavior.  
\\

\subsubsection{Virtual Sheep Allocator}\label{methodology:Virtualsheepallocator}
The virtual sheep allocator converts the set of frontiers,  $\mathcal{F}$, to a set of virtual sheep $\mathcal{V}$, represented as (position,weight) tuples:
\begin{equation*}
    \mathcal{V} (= [(v_1,w_1),(v_2,w_2),...,(v_{n_v},w_{n_v})])
\end{equation*}
%with enumerated .  
where $(n_v <n_f)$. Each robot $R_i$ downsamples the detected frontiers $f_k \in \mathcal{F}$, at a resolution, $f_{res}$. This downsampling mimics inter-agent repulsive force of the real-shepherding model, ensuring that no two virtual sheep are closer than $f_{res}$. %The number of unexplored cells in the square of size $f_{res}$, centered at every $v_i \in [v_1 \dots v_{n_v}]$ is mapped to the weight $w_i$ of the corresponding virtual sheep. 
The weight \( w_i \) of each virtual sheep \( v_i \) is determined by the number of unexplored cells within a square region of size \( f_{\text{res}} \), centered at \( v_i \).

\begin{figure}[h]
    \begin{subfigure}[t]{0.24\textwidth}
    \includegraphics[width=\textwidth]{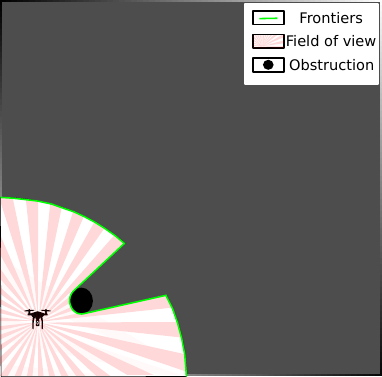}
    \captionsetup{justification=centering}
    \caption{Frontier Processor}
    \label{fig:methodology_frontier_processing}
  \end{subfigure}%\hfill
  \begin{subfigure}[t]{0.24\textwidth}
    \includegraphics[width=\textwidth]{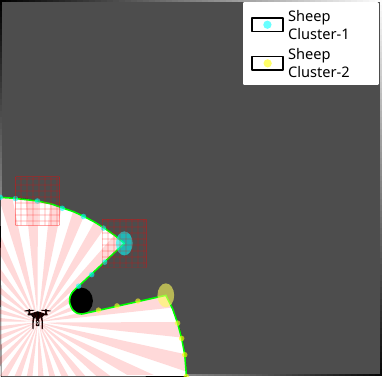}
    \captionsetup{justification=centering}
    \caption{Swarm Processor}
    \label{fig:methodology_swarm_processing}
  \end{subfigure}
  \caption{Virtual sheep are represented by blobs, with the radius depicting the weight of each virtual sheep. The sheep's weight, abstractly defining the exploration gain, is determined by the number of unexplored cells in the {red} grid centered at each sheep. Consequentially, heavier sheep are seen at the corners.}
  \label{fig:Frontier_and_Swarm_Processor}
\end{figure}

A simple $2$D representation of the resultant map frontiers for a single robot and the allocated virtual sheep is shown in Fig. \ref{fig:methodology_frontier_processing}  and Fig. \ref{fig:methodology_swarm_processing}, respectively.
\\
% \todo[inline]{Previously : The virtual sheep allocator
% converts the set of frontiers, F, to a set of nv (< nf )
% virtual sheep, V(= [(v1 , w1 ), (v2, w2 ), ..., (vnv , wnv )]) with enumerated (pose,weight) tuples. In this module, Ri initializes a virtual sheep of unit weight for each detected frontier, fk ∈ F. The swarm processor module uses a resolution, fresh , to discretize continuous frontiers and reduce the initialized virtual sheep count. The virtual sheep falling within fres of each other will be merged in terms of weight and moved to the center of mass of the sheep involved. This approach ensures that multiple frontiers within close proximity are weighed according to the exposed unexplored area of the combined frontiers. This also relates to inter-agent repulsive force [19], as no two virtual sheep can be closer than fres . A simple 2D representation of the resultant map frontiers for a single robot and the allocated virtual sheep is shown in Fig. 2a and Fig. 2b, respectively. }

% Reviewer 1: Unclear what this means. Maybe a diagram will help. If a continuous frontier is discretised, wouldn't there be a larger number of smaller frontiers? And hence more sheep? (one for each smaller frontier?)}

\subsubsection{Swarm Estimator}\label{methodology:Virtualsheepestimator}
%The swarm estimator estimates the behavior of $\mathcal{V}$ as $R_i$ moves through the terrain and generates the estimated swarm, $\hat{\mathcal{V}}$. This is done by modeling the forces of a sheep swarm with respect to the behavior of a frontier. The analogous forces are presented in \textbf{virtual sheep} column of Table \ref{tab:analogy_table} and explained below.

The swarm estimator predicts the movement of the virtual sheep swarm $\mathcal{V}$ as the robots navigate the environment, generating an estimated swarm $\hat{\mathcal{V}}$.  This estimation process models the forces influencing frontier behavior, enabling the robots to anticipate changes even when updates from the frontier processor are delayed.

The forces acting on each virtual sheep \( v_i \) are adapted from the original shepherding model. Table \ref{tab:analogy_table} presents the mapping of these forces to the frontier exploration context. Unlike real sheep, frontiers do not possess inertia, so the inertial force component is set to zero. The erroneous force, which introduces minor perturbations in the swarm, is retained to account for noise in frontier detection.

%Unlike sheep, the frontiers do not have an inertial component, wherein the frontiers move in the previously traveled direction. Thus, we are equating the analogous frontier inertial force to $0$. Erroneous force mimics the minor perturbations in the sheep swarm and is directly translated into our modeling.  

The sphere of influence of a robot $R_i$ is determined by its sensor range $L$. When a virtual sheep \( v_i \) is within the sensory range, it activates the clustering and predatory response of the virtual sheep. %The predatory response force is similar to  \cite{strombom2014solving}.
%The expected response of the sheep and frontiers, when in range of a predator/robot is to repel away. Along similar lines, we model the frontiers to disperse in the presence of a robot, thus the clustering force has a negative element to its shepherding counterpart. 
The predatory-response force causes frontiers to disperse away from the robot, similar to how sheep react to predators. Conversely, the clustering force behaves oppositely to its biological counterpart—rather than drawing sheep together, it encourages frontiers to spread out to reflect natural frontier expansion.

%This approach to modeling frontiers, $\mathcal{F}$, with a sheep swarm, $\mathcal{V}$, allows the subsequent shepherding module to heuristically estimate the next possible frontier by predicting the behavior of virtual sheep. When a new update of $\mathcal{F}$ is acquired from the frontier processor, $\mathcal{V}$ is updated and $\hat{\mathcal{V}}$ is reset to $\mathcal{V}$. The output of the swarm estimator, and eventually the swarm processor, is at a rate of $r_s >>r_f$ Hz. The increased rate allows the swarm processor to act as a buffer between the frontier and the predator processor modules. This ensures plausible delays from the frontier processor do not negatively impact the subsequent path planning. %Table \ref{tab:analogy_table} encapsulates the proposed analogy for modeling frontiers as a virtual sheep swarm. 

This approach allows robots to heuristically estimate the evolution of frontiers, ensuring smooth and adaptive exploration. When the frontier processor updates \( \mathcal{F} \), the estimated swarm \( \hat{\mathcal{V}} \) is reset to match the new frontier positions. The swarm estimator operates at a higher frequency \( r_s \gg r_f \), acting as a buffer between the frontier processor and the predator processor, ensuring delays do not negatively impact exploration efficiency.
\\

\subsubsection{Swarm Batching}\label{methodology:swarm_batching}
The estimated swarm $\hat{\mathcal{V}}$ can be discontinuous due to occlusions (Fig. \ref{fig:methodology_swarm_processing}) and the dispersive nature of the swarm estimator (Fig. \ref{fig:methodology_collecting2} and Fig. \ref{fig:methodology_herding2}). In such scenarios, treating the whole swarm as a single entity can lead to %preventable inter-robot collision, chasing similar frontiers, and repeated path planning.
unnecessary inter-robot collisions, redundant exploration of the same frontiers, and increased computational overhead from repeated path planning. To mitigate this, the swarm batching module clusters $\hat{\mathcal{V}}$ into $n_b$ distinct batches using a clustering algorithm \cite{kodinariya2013review}. 
Each batch is represented as a tuple ($v_b$,$w_b$), where $v_b$ is the center of mass of a cluster of $\hat{\mathcal{V}}$, and $w_b$ is the corresponding cumulative weight of the cluster. The set of virtual sheep swarm cluster descriptors, $\mathcal{V}_b = [(v_{b1},w_{b1}),(v_{b2},w_{b2}),...,(v_{b{n_b}},w_{b{n_b}})]$, is shared among robots $\mathcal{R}$. 

Each tuple in $\mathcal{V}_b$ corresponds to a unique cluster of virtual sheep in $\hat{\mathcal{V}}$, ensuring that each robot is assigned to a distinct frontier region. %or $\hat{\mathcal{V}}(\mathcal{V}_b(k))$ represents a unique sheep cluster, $\forall k \in [1,n_b]$. 
%Exploring the batch of virtual sheep with the maximum cumulative weight or maximum exploration gain. 
The assigned cluster for each robot $R_i$ is determined by selecting the batch that maximizes exploration gain while minimizing redundant coverage.

To achieve this, a distance penalty is introduced to encourage each robot to prioritize closer clusters, preventing unnecessary switching between batches or multiple robots converging on the same frontier. Each robot computes the distance to each batch centroid $v_b$ and normalizes it against the maximum observed distance $d_{max}$, while the weight is normalized against the maximum weight $w_{max}$. The selection of the optimal cluster $\hat{\mathcal{V}_s}$ for each robot is formulated as:
\begin{equation} \label{eq:min_dist_mass}
\underset{\forall i \in [1,n_r]}{B_i} = \arg\max_{(v_{b},w_{b}) \in \mathcal{V}_b} \left( \lambda_m \frac{w_b}{w_{max}} - \lambda_d \frac{\| \bar{R_i} - v_{b} \|_2}{d_{max}}\right) 
\end{equation}
where $\lambda_m$ and $\lambda_d$ are tunable parameters controlling the trade-off between exploration gain and travel distance. For UAVs, a lower $\lambda_d$ minimizes the impact of distance in the selection process, while for UGVs, increasing $\lambda_d$ helps prioritize nearby clusters to optimize traversal efficiency.

This batch-based allocation strategy targets efficient frontier distribution among agents, reduces redundant exploration, and enhances overall system robustness.

%We incorporate a distance penalty to ensure each agent prefers a closer swarm batch to prevent unwarranted switching between batches and multiple robots chasing the same frontiers. 
%An allocated batch is removed from the poll and updated across the team, thus enforcing a one-to-one swarm batch to robot mapping. 
%Each robot within proximity calculates the distance to each $v_b$ and the corresponding weight. Normalization with respect to the maximum of both distance, $d_{max}$ and weight, $w_{max}$ is carried out to bound the values. Each robot, $R_i$ chooses the cluster, $\hat{\mathcal{V}_s} = \hat{\mathcal{V}}(\mathcal{V}_b(B_i))$, where $B_i$ maximizes Eq. (\ref{eq:min_dist_mass}). The distance coefficient $\lambda_d$ and the weight coefficient $\lambda_m$ are tunable parameters that can be tuned according to the robotic team involved. For UAVs, a lower $\lambda_d$ promises a reduced impact on distance in the maximization problem. For UGVs, traversing longer distances can be more taxing and thus increasing $\lambda_d$ will favour nearby heavier batches. 

%\begin{equation} \label{eq:min_dist_mass}
%\underset{\forall i \in [1,n_r]}{B_i} = %\arg\max_{(v_{b},w_{b}) \in \mathcal{V}_b} \left( %\lambda_m \frac{m_b}{w_{max}} - \lambda_d \frac{\| %\bar{R_i} - v_{b} \|_2}{d_{max}}\right) 
%\end{equation}

\subsection{Predator Processor}\label{sec:shepherding}

The predator processor module governs the robots' behavior in controlling the assigned subset of the virtual sheep swarm $\hat{\mathcal{V}_{s}}$ to maximize exploration efficiency. In the real shepherding model, a heuristic predator operates in two modes: collecting and herding, depending on the compactness of the sheep swarm. In the FroShe framework, a batch of virtual sheep $\hat{\mathcal{V}_{s}}$ is considered compact if all its members are within a threshold distance $d_t$ from its center of mass $C_m$. The predator processor determines the appropriate mode based on this compactness criterion and adjusts the robot’s behavior accordingly.

%The heuristic predator controls the selected subset of the virtual sheep swarm, $\hat{\mathcal{V}_{s}}$, identified by $R_i$'s swarm processor module (Sec.\ref{methodology:swarmprocessor}), to maximize exploration gain. In \cite{strombom2014solving}, a heuristic predator has two modes of operation: \textit{collecting} and \textit{herding} depending on the compactness of the sheep swarm. $\hat{\mathcal{V}_{s}}$ is deemed to be compact if all the sheep in $\hat{\mathcal{V}_{s}}$ is within a distance threshold, $d_t$ from the center of mass of $\hat{\mathcal{V}_{s}}$, $C_m$. A global perception parameter, $p_p$, determines how closely $R_i$ will approach to influence $\hat{\mathcal{V}_{s}}$. As the robot moves through the terrain, the estimates $\hat{\mathcal{V}}$ and $\hat{\mathcal{V}_s}$ change in congruence with the forces of Table \ref{tab:analogy_table}.

\subsubsection{Collecting Mode}\label{subsec:collecting}
\begin{figure}[h]
  \centering
  \begin{subfigure}[t]{0.16\textwidth}
    \centering
    \includegraphics[width=\textwidth]{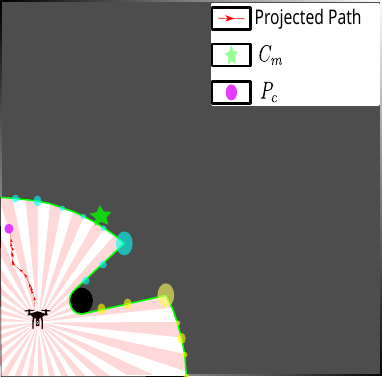}
    \captionsetup{justification=centering}
    \caption{}
    \label{fig:methodology_collecting1}
  \end{subfigure}%
  \hfill
  \begin{subfigure}[t]{0.16\textwidth}
    \centering
    \includegraphics[width=\textwidth]{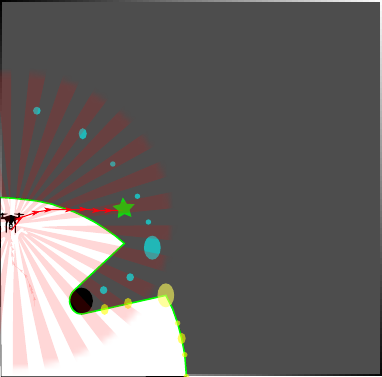}
    \captionsetup{justification=centering}
    \caption{}
    \label{fig:methodology_collecting2}
  \end{subfigure}%
  \hfill
  \begin{subfigure}[t]{0.16\textwidth}
    \centering
    \includegraphics[width=\textwidth]{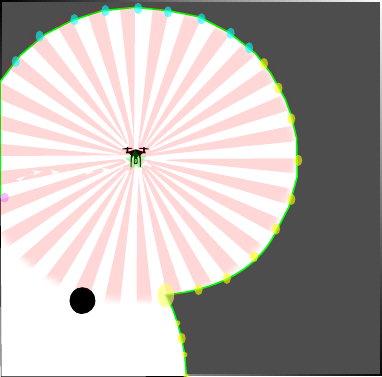}
    \captionsetup{justification=centering}
    \caption{}
    \label{fig:methodology_collecting3}
  \end{subfigure}
  \caption{Collecting mode. (a) Planned trajectory to $P_c$ (b) Planned trajectory from $P_c$ to $C_m$ (c) Final $\mathcal{F}$ in $\mathcal{M}$ after collecting.}
  \label{fig:Collecting_Methodology}
\end{figure}

When the swarm is not compact, the robot prioritizes pushing outlier virtual sheep toward the cluster’s center of mass $C_m$ to maintain coherence. A collecting pose $P_c$ is generated near the furthest virtual sheep ${v}_{f}$ in $\hat{\mathcal{V}_{s}}$, guiding the robot to reposition itself strategically. The collecting pose is computed as:
\begin{equation}\label{eqn:collecting}
P_c = v_{f} + p_p * L *  \frac{C_m-v_{f}}{||C_m-v_{f}||}
\end{equation}
where $p_p$ is a global perception parameter that determines how closely the robot approaches the swarm, and 
$L$ is the robot’s perception range. Once positioned at 
$P_c$, the robot moves toward $C_m$ effectively pushing the scattered virtual sheep to the center (Fig. \ref{fig:methodology_collecting2}).

%When the swarm is not compact, a predator attempts to push the outlier sheep towards the center of mass, to keep the swarm compact. Analogously, a collecting pose $P_c$, formulated in Eq. (\ref{eqn:collecting}), is generated close to the furthest sheep, ${v}_{f}$, in $\hat{\mathcal{V}_{s}}$. A trajectory is planned to $P_c$ (Fig. \ref{fig:methodology_collecting1}) and from $P_c$ to $C_m$(Fig. \ref{fig:methodology_collecting2}).  
\subsubsection{Herding}\label{subsec:herding}
When the swarm is compact, the robot transitions to herding mode, pushing the swarm toward a new unexplored frontier region. Unlike conventional shepherding, where the destination is predefined, exploration tasks lack a fixed target. Instead, the robot generates a driving pose $P_d$, which encourages movement toward the next area of high exploration gain. This is determined by identifying the center of mass $C_h$  of the adjacent heaviest swarm batch and computing $P_d$ as:

\begin{equation}\label{eqn:herding}
P_d = C_{m} - p_p * L *  \frac{C_m-C_h}{||C_m-C_h||}
\end{equation}

\begin{figure}[h]
  \centering
  \begin{subfigure}[t]{0.16\textwidth}
    \centering
    \includegraphics[width=\textwidth]{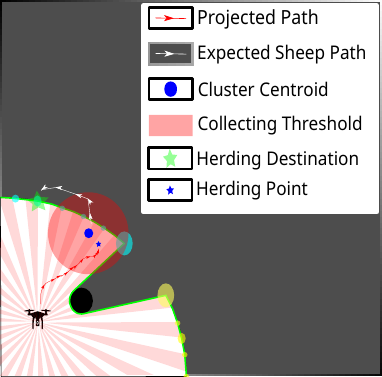}
    \captionsetup{justification=centering}
    \caption{}
    \label{fig:methodology_herding1}
  \end{subfigure}%
  \hfill
  \begin{subfigure}[t]{0.16\textwidth}
    \centering
    \includegraphics[width=\textwidth]{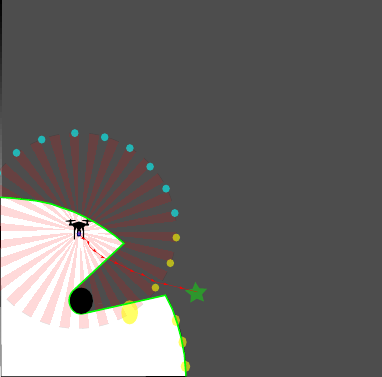}
    \captionsetup{justification=centering}
    \caption{}
    \label{fig:methodology_herding2}
  \end{subfigure}%
  \hfill
  \begin{subfigure}[t]{0.16\textwidth}
    \centering
    \includegraphics[width=\textwidth]{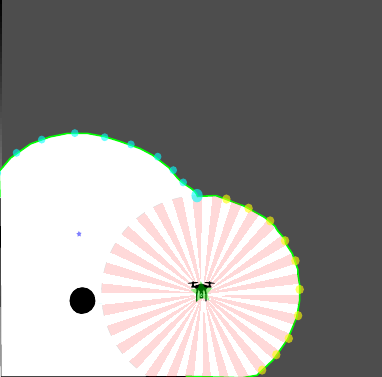}
    \captionsetup{justification=centering}
    \caption{}
    \label{fig:methodology_herding3}
  \end{subfigure}
  \caption{Herding mode (a) Planned trajectory to $P_d$ (b) Planned trajectory from $P_d$ to $C_h$ (c) Final $\mathcal{F}$ in $\mathcal{M}$ after herding.}
  \label{fig:Herding_Methodology}
\end{figure}

A trajectory is then planned from the current position to $P_d$ (Fig. \ref{fig:methodology_herding1}) and subsequently from $P_d$ to to $C_h$ (Fig. \ref{fig:methodology_herding2}), effectively pushing the swarm toward high-priority frontier regions. Inherently, the robot attempts to merge the two batches with the largest possible exploratory gain.

%When the swarm is compact, the heuristic predator attempts to push the swarm to the desired destination. Since an exploration task is devoid of a destination, a driving pose $P_d$, as formulated in Eq. (\ref{eqn:herding}), is generated with respect to $C_m$ of the current frontier batch and the center of mass of adjacent heaviest swarm batch, $C_h$. Inherently, the predator attempts to merge the two batches with the largest possible exploratory gain. $P_d$ is formulated in Eq. (\ref{eqn:herding}), and a trajectory is planned to $P_d$ (Fig. \ref{fig:methodology_herding1}) and from $P_d$ to $C_h$ (Fig. \ref{fig:methodology_herding2}). 

Fig. \ref{fig:methodology_collecting2} and Fig. \ref{fig:methodology_herding2} depicts the missing map update from the frontier processor. The figures also portray the effect of the swarm estimator in the absence of a map update.
% \begin{figure}[h]
%   \begin{subfigure}[t]{0.25\textwidth}
%     \includegraphics[width=\textwidth]{Figures/Shepherd_Herding.pdf}
%     \captionsetup{justification=centering}
%     \caption{}
%     \label{fig:methodology_herding}
%   \end{subfigure}%\hfill
%   \begin{subfigure}[t]{0.25\textwidth}
%     \includegraphics[width=\textwidth]{Figures/Shepherd_Herding_t2.pdf}
%     \captionsetup{justification=centering}
%     \caption{}
%     \label{fig:methodology_herding_t1}
%   \end{subfigure}
%   \caption{ Herding mode\checkplease{Inkscape issue, three images here}} 
%   \label{fig:Herding_Methodology}
% \end{figure}

\subsubsection{Exploration Rate Monitor}\label{subsec:expratemonitor}

To regulate the transition between collecting and herding, we introduce an exploration rate monitor that analyzes the rate of change in explored area $\Delta E$ using a time-series moving average model. The moving average at time step $n$ is defined as:
\begin{equation}\label{eqn:movingaverage}
MA(n) = \frac{1}{n} \sum_{i=0}^{n-1} \Delta E_{t-i}
\end{equation}
where $MA(n)$ captures exploration trends over time. A fast-moving average (FMA) with a short time window ($\sim50$ iterations) tracks recent performance, while a slow-moving average (SMA) with a longer window ($\sim200$ iterations) represents long-term expectations.

\begin{figure}[t]
    \includegraphics[height=0.2\textwidth,width=0.48\textwidth]{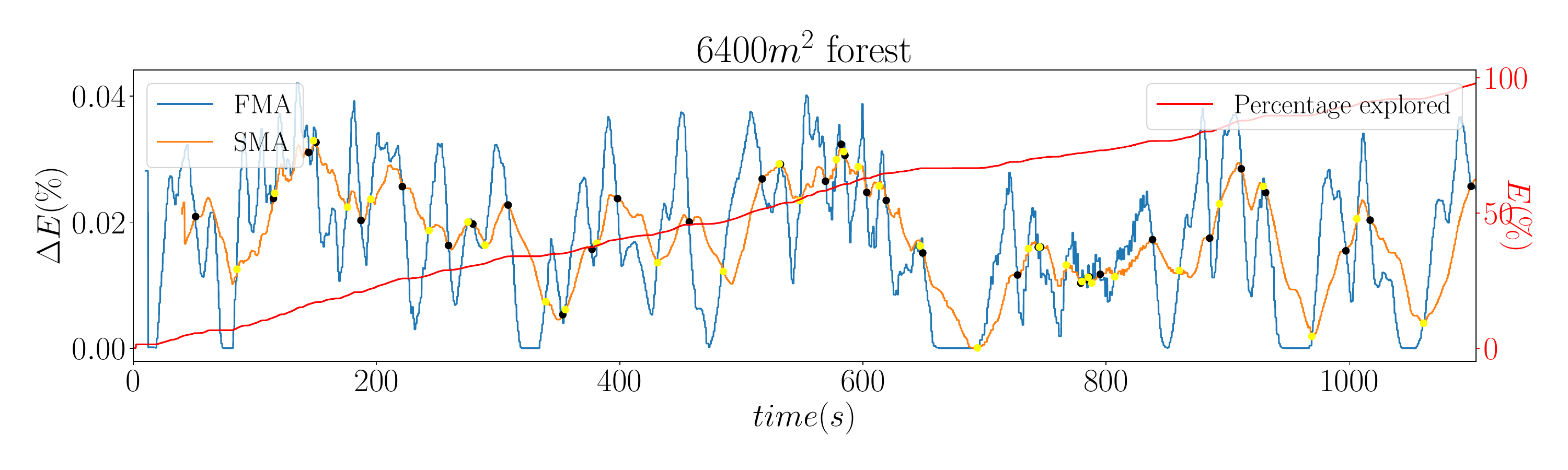}
    %\captionsetup{justification=centering}
    \caption{Exploration rate monitoring in a single-agent exploration scenario within a $6400m^2$ forest-like environment. The plot tracks the percentage of explored area $E$ over time and its rate of change $\Delta E$. Black dots indicate mode switches triggered by changes in $d_t$, while yellow dots mark instances where the current mode is retained. The Fast Moving Average (FMA) and Slow Moving Average (SMA) guide these transitions, ensuring steady exploration progress.}

    \label{fig:moving_average}
\end{figure}

Mode switching is triggered when the FMA drops below the SMA, indicating a slowdown in exploration progress. If the current dominant mode is collecting, the threshold $d_{t}$ is increased to encourage a transition to herding. Conversely, if the dominant mode is herding, decreasing 
$d_{t}$ promotes collecting. When SMA exceeds FMA, the strategy remains unchanged, ensuring stability in decision-making. Fig. \ref{fig:moving_average}, illustrates how this mechanism dynamically adapts the robot behavior to optimize exploration.

This adaptive switching mechanism ensures that robots efficiently balance between consolidating frontiers and pushing exploration boundaries, maximizing coverage while minimizing redundant movements.

\section{RESULTS } 

\label{sec:results}
%Section \ref{Results:Simulation}    provides an in-depth simulation analysis, carried out to compare the FroShe with SOTA multi-agent exploration strategies. Details of an implementation of the proposed algorithm in a real-world a forest-like scenario is showcased in Section \ref{Results:Real}. 
\subsection{Simulation Results} \label{Results:Simulation}
\subsubsection{Setup}

To evaluate the performance of FroShe, we conducted extensive simulations using the MRS UAV system \cite{baca2021mrs} in ROS Noetic. This framework provides a comprehensive multi-UAV simulation environment including path planning with collision avoidance, realistic sensor integration, and various 3D mapping algorithms \cite{zhang2014loam,hornung2013octomap}.
Additionally, the simulator incorporates accurate sensor and actuator models, enabling realistic Software-In-The-Loop (SITL) simulations that closely mirror real-world conditions.
%are carried out in the MRS UAV system  \cite{baca2021mrs} on ROS Noetic, which provides an end-to-end multi-UAV framework complete with path planning enabled with collision avoidance and realistic-sensor integration along with various $3$-D mapping algorithms  \cite{zhang2014loam,hornung2013octomap}. 
%Additionally,   \cite{baca2021mrs} replicates realistic simulations that can be transferred to real-world implementation in a multi-UAV system. 

\begin{figure}[h]
  \begin{subfigure}[t]{0.25\textwidth}
    \includegraphics[width=\textwidth]{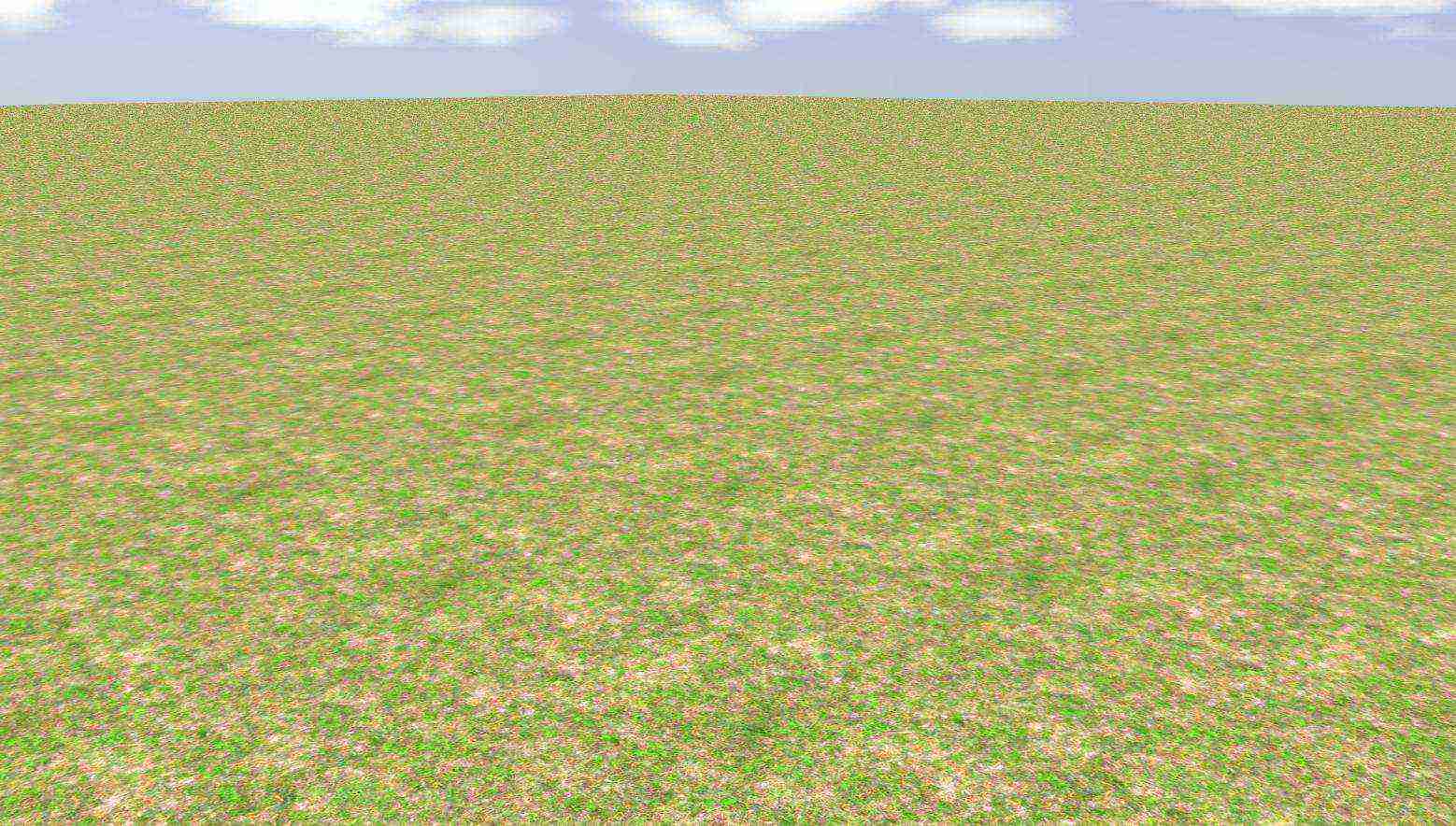}
    \captionsetup{justification=centering}
    \caption{}
    \label{fig:results_simulation_environment_grass_plane}
  \end{subfigure}%\hfill
  \begin{subfigure}[t]{0.25\textwidth}
    \includegraphics[width=\textwidth]{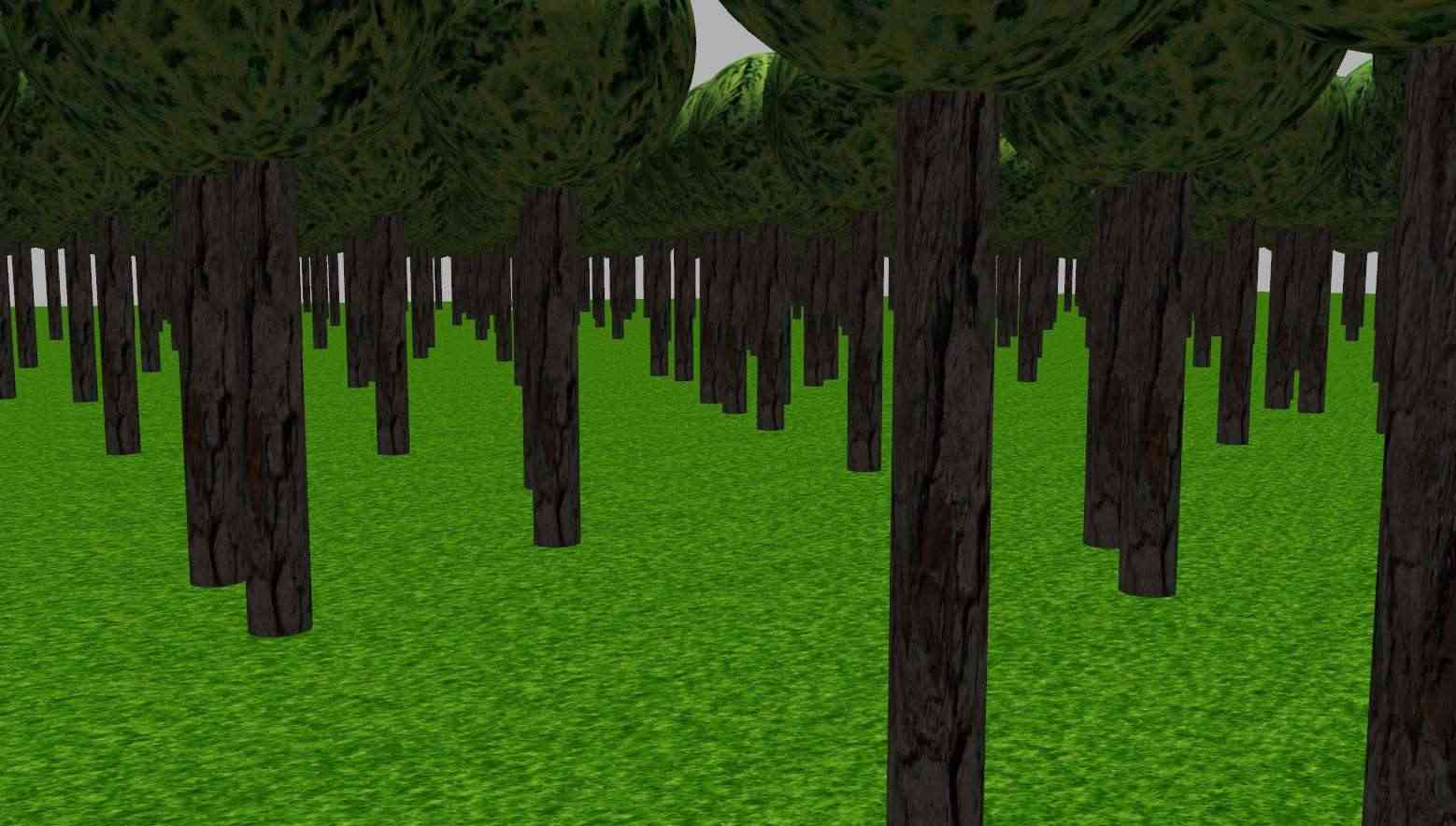}
    \captionsetup{justification=centering}
    \caption{}
    \label{fig:results_simulation_environment_forest}
  \end{subfigure}
  % \caption{ Simulation Environments. The grass plane (a), is devoid of obstacles and helps analyze the exploration strategies with straight-forward path planning with no obstacle avoidance. In contrast, forest (b) provides a cluttered environment, with an average tree density of 0.05 trees/$m^2$.} 
  \caption{ Simulation Environments. (a) Grass plane: This environment is devoid of obstacles, allowing for straightforward path planning without obstacle avoidance. (b) Forest: This cluttered environment has an average tree density of 0.05 trees/m², providing a more challenging scenario for testing exploration strategies.} 
  \label{fig:Experimentalsetup_simulation}
\end{figure}

We tested FroShe in two distinct environments with different obstacle densities, as shown in Fig. \ref{fig:Experimentalsetup_simulation}. Each environment was subdivided into square areas of $1600$m$^2$, $3600$m$^2$, and $6400$m$^2$ allowing us to analyze performance variations across different exploration scales. The maximum exploration time was capped at $3600$ seconds, and each experiment was repeated $25$ times, with UAVs initialized within a $5$ m radius of one another at a designated starting point.

%square patches for exploration in each environment to analyze the effect of varying areas of interest. The experiment is considered a failure if exploration is not completed within a predefined duration of $3600$s. Results are analyzed across $25$ runs with the drones spawned within $5$m of each other, in one corner of the area of interest.

For simulation, we modeled the agents as f$550$ UAVs equipped with an Ouster (OS$1$-$128$) LIDAR, with perception range limited to $10$m. The maximum horizontal speed, maximum vertical speed, maximum acceleration and maximum jerk were set to ($4$m/s, $2$m/s, $2$m/s$^2$, $40$m/s$^3$) for grass plane and ($1$m/s, $1$m/s, $1$m/s$^2$, $20$m/s$^3$) for forest, to ensure safety. The maximum allowed flight height was limited to $4$m  and the minimum was bounded at $1$m. We utilize octomapping  \cite{hornung2013octomap} to generate the $3$-D map of the environment. The resultant local occupancy grid is globally shared across other UAVs.

To benchmark FroShe’s performance, we compared it against FAME \cite{bartolomei2023fast} and the exploration strategy proposed by Burgard et al, \cite{burgard2005coordinated}. 
FAME \footnote{\text{https://github.com/VIS4ROB-lab/fast\_multi\_robot\_exploration}} employs a dual-mode approach incorporating a traveling salesman-based optimization, originally designed for depth cameras with a limited field of view. For a fair comparison, we adapted FAME\footnote{\text{https://github.com/johnd2010/fast\_multi\_robot\_exploration}} to utilize a $360^\circ$ LiDAR. The approach by Burgard et al. prioritizes exploration based on a utility-value function, continuously updated along planned paths. Target poses from both methods were integrated into the MRS UAV framework for trajectory execution with collision avoidance. No communication constraints were imposed in these simulations.

%FAME provides a dual-mode approach to exploration incorporating a traveling salesman-based optimization. The algorithm, FAME\footnote{\text{https://github.com/VIS4ROB-lab/fast\_multi\_robot\_exploration}}, originally developed for a depth camera with a limited field of view, has been incorporated to include a $360^\circ$ LIDAR.  Burgard et al. approach explores with a utility-value function updated along the path to the target poses. The generated target poses are shared to MRS UAV framework\cite{baca2021mrs}, to generate the trajectory after incorporating collision avoidance. No communication constraints are imposed on the agents.%The algorithm, FAME\footnote{\text{https://github.com/VIS4ROB-lab/fast\_multi\_robot\_exploration}}, originally developed for a depth camera with a limited field of view, has been incorporated to include a $360$ LIDAR\footnote{\text{https://github.com/johnd2010/fast\_multi\_robot\_exploration}} in   \cite{baca2021mrs}.   \cite{burgard2005coordinated} explores with a utility-value function updated along the path to the target poses. The generated target poses are shared to   \cite{baca2021mrs}, to generate the trajectory after incorporating collision avoidance. No communication constraints are imposed on the agents.

\subsubsection{Analysis}

\begin{figure*}
 \centering
 \begin{minipage}[h]{\textwidth}
     \centering
     \includegraphics[width=\textwidth]{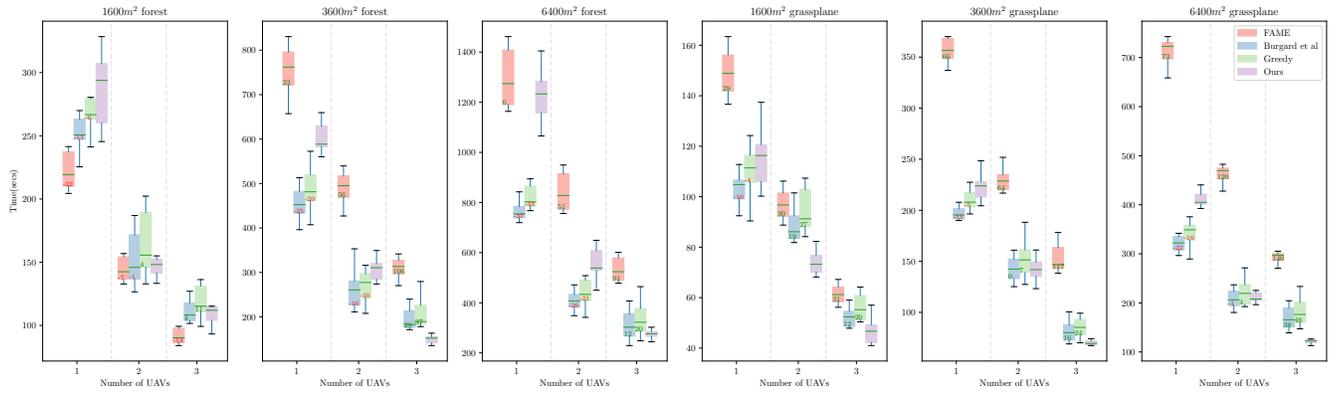}
    \caption{Time-taken analysis for varying number of agents and varying area-environments for different exploration strategies. The numbers within the bars depict the percentage change with respect to the time taken by FroShe, with an increase or decrease in performance colored accordingly. Due to varying time taken across each subplot, it is to be noted that the time axis (in seconds) is not shared across the subplots.}
     \label{fig:simulation_time_analysis}
 \end{minipage}%

 \end{figure*}

Fig. \ref{fig:simulation_time_analysis} presents box plots illustrating exploration times for different numbers of agents $(1$ to $3)$ across various environments and area sizes. Each subplot compares FroShe, Burgard et al., Greedy exploration, and FAME. Notably, FAME's performance deteriorates compared to its reported results in \cite{bartolomei2023fast}, likely due to the transition from a limited field-of-view depth camera to a $360^\circ$ LiDAR, as well as the shift to a Gazebo-based Software-In-The-Loop (SITL) simulation. The latter incorporates accurate sensor and actuator models, realistic physics, and sensor noise, making the evaluation more representative of real-world conditions. While a detailed ablation study comparing FAME and FroShe could provide deeper insights, such an analysis is beyond the scope of this work. Furthermore,  additional agents could further improve efficiency, our experiments were limited to $3$ UAVs due to the computational demands of the near-realistic SITL simulator, which fully models physics, sensors, and actuators.

%Fig. \ref{fig:simulation_time_analysis} shows the time box plots for a varying number of agents $(1$ to $3)$. Each subplot depicts a different combination of environment and area, with each bar plot representing FroShe,  Burgard et al., Greedy exploration and  FAME. FAME's notable inconsistent performance deterioration with respect to what is reported in  \cite{bartolomei2023fast} could be attributed to the switch to the $360^\circ$  LiDAR and switch to Gazebo simulator with Software-In-The-Loop (SITL) simulation, accurate sensor and actuator models, as well as realistic physics simulator and sensor noises. An ablation study of   \cite{bartolomei2023fast} with respect to FroShe will provide further insights but is beyond the current article's scope.

\begin{figure}[t]
    \includegraphics[width=0.5\textwidth]{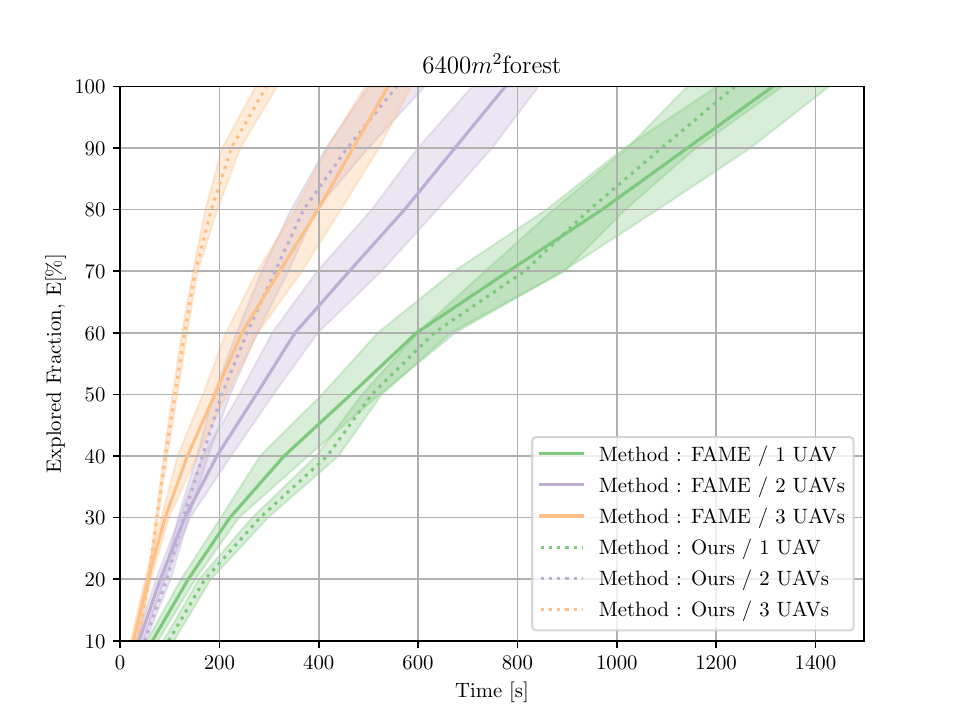}
    % \captionsetup{justification=centering}
    \caption{Exploration rates over different numbers of UAVs for FroShe and  FAME \cite{bartolomei2023fast} in an $6400m^2$ forest.}
    \label{fig:Results_exploration_time}
\end{figure}

As expected (Fig. \ref{fig:simulation_time_analysis}, Fig. \ref{fig:Results_exploration_time}), total exploration time decreases as the number of agents increases. However, the reduction in variance is particularly evident in FroShe, demonstrating its ability to maintain consistent exploration times regardless of environmental complexity. This robustness across different scenarios suggests that FroShe can reliably achieve full coverage within a predictable timeframe, making it more adaptable than other approaches.

In single-agent scenarios, both Burgard et al. and Greedy strategies outperform FroShe, due to FroShe’s continuous switching between herding and collecting behaviors, which increases travel distance per unit of exploration gain. However, with multiple agents, FroShe benefits from its swarm allocation mechanism (Sec \ref{sec:frontierprocessing}, Sec \ref{methodology:swarm_batching}), which effectively segments the exploration space and reduces redundant travel. This improves overall efficiency, leading to a $25\%$  reduction in exploration time compared to other methods when using three agents, across all tested environments.
%As expected and denoted in Fig. \ref{fig:simulation_time_analysis} and Fig. \ref{fig:Results_exploration_time}, there is a consistent improvement in total exploration time with the number of agents. However, the decrease in variance with the number of agents is more evident in FroShe, demonstrating that the proposed algorithm consistently covers the area within a given time. This consistency across all areas and environments demonstrates the robustness of the proposed algorithm compared to others.
%For single-agent scenarios, both  Burgard et al. and greedy strategies outperform frontier shepherding. The continuous switching between herding and collecting in a single agent scenario makes the proposed algorithm more time-consuming as more distance is to be covered for reduced exploration gain. However, as the number of agents increases, FroShe benefits from the initial delegation of segregated frontier clusters (Sec \ref{sec:frontierprocessing}) via swarm allocation (Sec \ref{methodology:swarm_batching}). This reduces the expected distance travelled while switching between shepherding modes, thereby improving time taken. With $3$ agents, we can see an average improvement of $25\%$ in time-taken, with respect to the other approaches for all areas across all environments.
A key advantage of FroShe is its robust performance across different environments and exploration scales without requiring extensive parameter tuning. This was validated by maintaining a single set of parameters, optimized for a $1600m^2$ forest scenario, across all simulation experiments. 
%FroShe also demonstrates a clear improvement in exploration efficiency as coverage area increases. Specifically, for a three-agent scenario, FroShe achieves a minimum of $12\%$ improvement in forest environments and up to $48\%$ in open grassland compared to other approaches. 
%Further analysis of the $6400 m^2$ forest scenario (Fig. \ref{fig:Results_exploration_time}) provides a direct comparison between FroShe and FAME. While both methods exhibit a steady decline in total exploration time as the number of agents increases, FroShe consistently achieves lower variance, reinforcing its adaptability and reliability in large-scale exploration tasks.

%A prime advantage of FroShe over other algorithms is the robustness across varying environments and exploration areas without the need for parametric tuning. This is validated by maintaining a constant set of parameters that are optimized for a $1600m^2$ forest scenario throughout all simulations.
% \todo[inline]{Previous: "It is also essential to point out that FAME performs best in $1600$m$^2$ forest, which could be attributed to the fact that the initial algorithm was tuned to suit a $2500$m$^2$ forest. To check the robustness and flexibility of each algorithm with minimal intervention, no parameters were changed throughout the simulation analysis." 

% Reviewer 1: So is this an unfair comparison? Should it be retuned? How much of an improvement is expected after doing so? }

FroShe also show a clear improvement in the exploration time as the coverage area increases, with a minimum of $12\%$ in forest and $48\%$ in grass plane for $3$ agent scenario. A closer analysis of exploration time (Fig. \ref{fig:Results_exploration_time}) is  carried out on the $6400 m^2$ forest scenario with both FroShe and FAME. %As a state-of-the-art forest exploration algorithm, FAME was designed to navigate forested environments and efficiently capture trails. 
While both methods exhibit a consistent decline in total exploration time, the reduced variance showcases the robustness of FroShe across different conditions.

\begin{figure}[t]
  \begin{subfigure}[t]{0.25\textwidth}
    \includegraphics[width=\columnwidth]{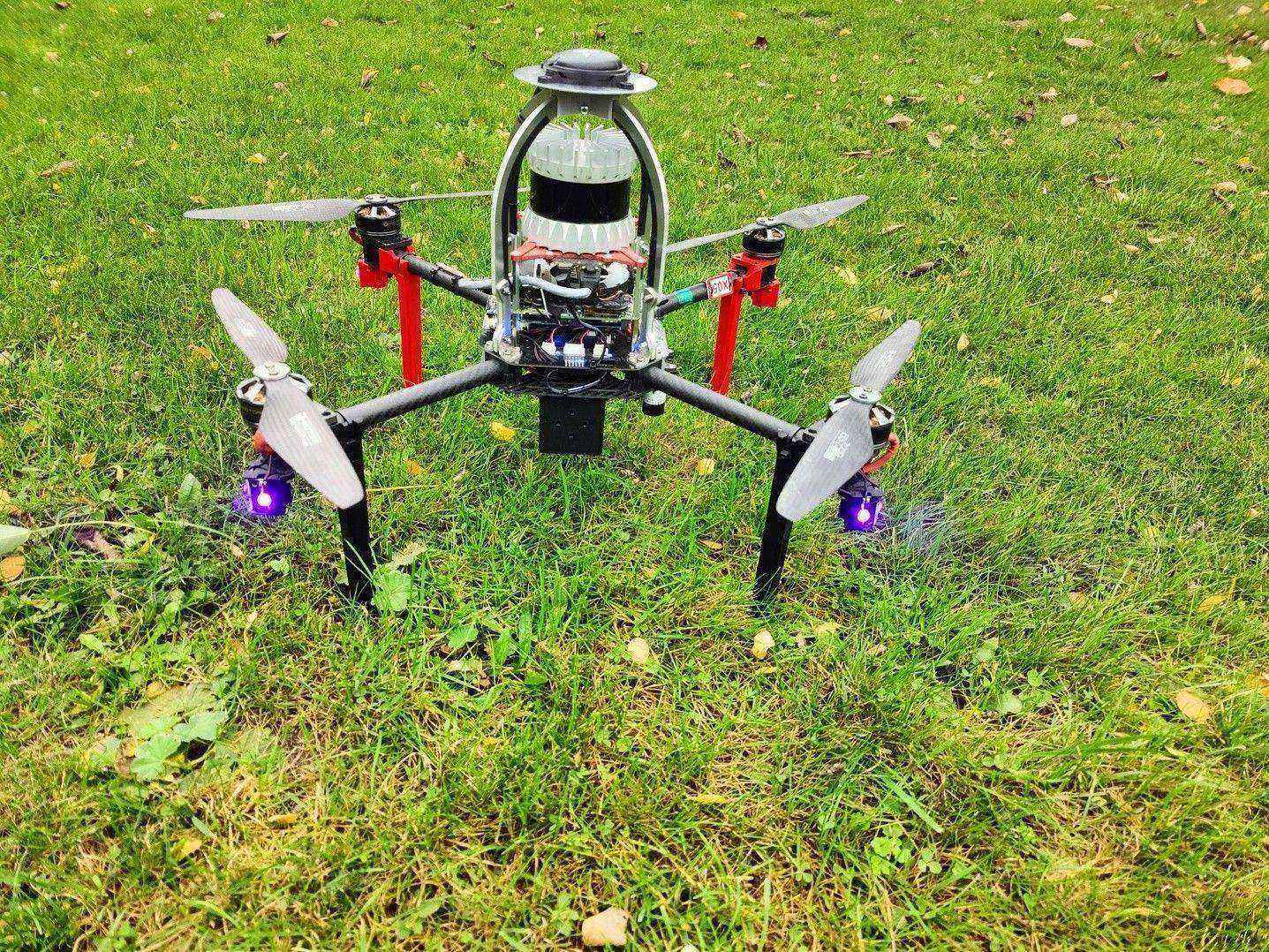}
    \captionsetup{justification=centering}
    \caption{}
    \label{fig:results_real_drone}
  \end{subfigure}%\hfill
  \begin{subfigure}[t]{0.25\textwidth}
    \includegraphics[width=\columnwidth,height=0.75\textwidth]{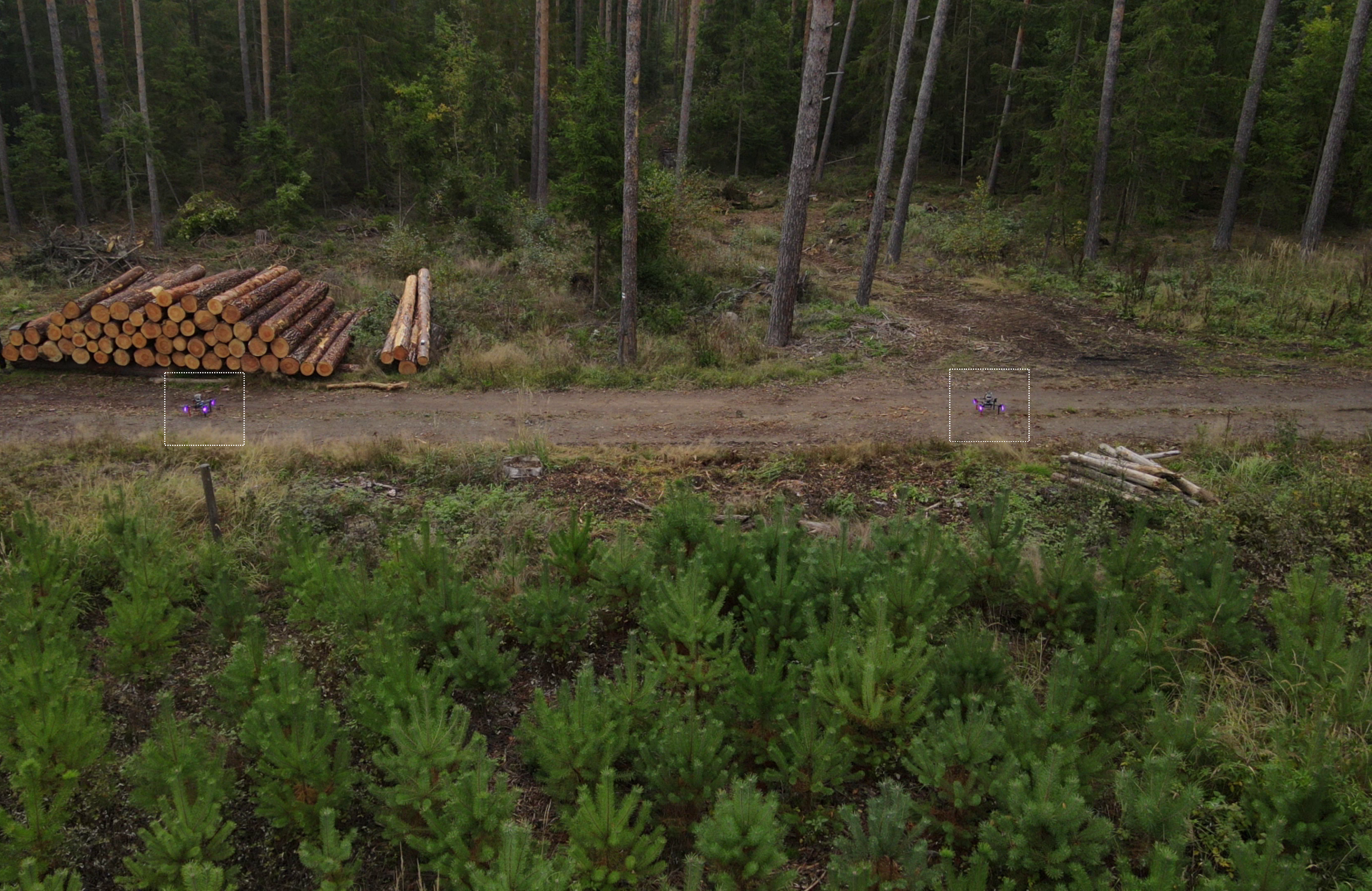}
    \captionsetup{justification=centering}
    \caption{}
    \label{fig:results_real_setup}
  \end{subfigure}%\hfill
  \caption{ Real World Experiments with $2$ X$500$ drones. The drones were separated by $10$m. } 
  \label{fig:Experimentalsetup_real}
\end{figure}

\begin{figure}[t]
    \includegraphics[width=0.5\textwidth]{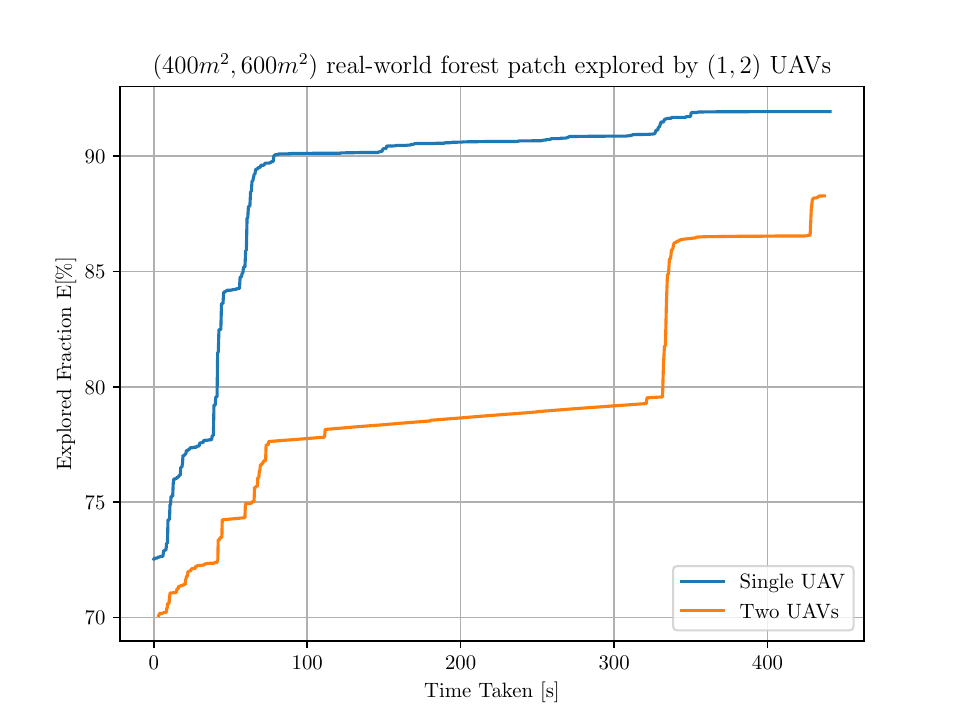}
    % \captionsetup{justification=centering}
    \caption{Exploration rates over different numbers of UAVs exploring a real-world forest-like environment with FroShe. It is to be noted that for a $2$ UAV scenario, the area of interest is $600$m$^2$ and $400$m$^2$ for $1$ UAV scenario.}
    \label{fig:Real_Exploration_Time}
\end{figure}

\subsection{Real World Experiments}\label{Results:Real}

To validate the performance of FroShe in real-world conditions, we conducted experiments with the help of the AeroSTREAM Open Remote Laboratory\footnote{https://fly4future.com/aerostream-open-remote-laboratory/}. The tests were performed in a forest-like environment, as shown in Fig. \ref{fig:results_real_setup}, using two X500 drones Fig. \ref{fig:results_real_drone}, each equipped with an Ouster OS1 LiDAR for perception and mapping. The objective was to evaluate FroShe’s adaptability to real-world challenges, such as sensor noise and  environmental occlusions. 

We conducted two sets of experiments: a single-UAV exploration over a $400$m$^2$ area and a dual-UAV exploration over a $600$m$^2$ area. The LiDAR sensing range was limited to 10 meters, similar to the constraints used in simulation, and to allow exploration in the limited space available for the experiments. Fig. \ref{fig:Real_Exploration_Time} presents the exploration progress for both scenarios, showing a steady increase in explored area over time.
The results confirm FroShe’s ability to efficiently coordinate multiple agents in an unstructured, real-world environment. The algorithm successfully allocated frontiers between UAVs, minimizing redundant exploration while ensuring complete coverage. The dual-UAV scenario exhibited a clear improvement in exploration speed, demonstrating the framework’s scalability. These findings reinforce FroShe’s effectiveness in real-world deployment, highlighting its potential for large-scale autonomous exploration tasks.

%We leveraged the seamless integration of the MRS UAV framework \cite{baca2021mrs} onto a real-world scenario in a forest-like environment, pictured in Fig. \ref{fig:results_real_setup}.
%The experiments involved two X500 drones, shown in Fig. \ref{fig:results_real_drone}, each equipped with an Ouster OS1 LiDAR. The experiments were conducted in a forest-like environment to generate a scenario that demands obstacle avoidance. Single UAV and dual UAV exploration experiments were conducted, with the UAV(s) tasked to explore a $400$m$^2$ and $600$m$^2$ area, with the lidar range clipped at $10$m. The exploration trend for both experiments is showcased in Fig. \ref{fig:Real_Exploration_Time}. 

%%%%%%%%%%%%%%%%%%%%%%%%%%%%%%%%%%%%%%%%%%%%%%%%%%%%%%%%%%%%%%%%%%%%%%%%%%%%%%%%

\section{CONCLUSION}

\label{sec:conclusion}
This paper introduced Frontier Shepherding (FroShe), a bio-inspired multi-robot framework for large-scale exploration. By modeling frontiers as a virtual sheep swarm and using a shepherding-based heuristic, FroShe enables efficient and scalable exploration across diverse environments. The framework is modular, online, and decentralized, allowing for rapid deployment with minimal parameter tuning, making it well-suited for computationally constrained robots operating in unknown and hazardous terrains.

Extensive simulations in forest and grassland environments demonstrated that FroShe outperforms state-of-the-art exploration strategies as the number of agents increases. With three UAVs, FroShe achieved an average $25\%$ reduction in exploration time compared to other methods, while also maintaining greater consistency across different environment sizes and obstacle densities. Real-world experiments in a forest-like environment further validated its robustness, confirming its ability to efficiently coordinate multiple agents and adapt to unstructured, real-world conditions.

For future work, we aim to extend FroShe to heterogeneous robot teams, leveraging agents with different sensing and mobility capabilities. Additionally, integrating a dual-velocity \cite{bartolomei2023fast} could further optimize performance, especially in heterogeneous teams. Another avenue for improvement is refining the exploration rate monitor to operate at a global level, allowing for more coordinated mode switching between herding and collecting based on team-wide performance. These enhancements will further improve FroShe’s adaptability and scalability for real-world autonomous exploration tasks.

\section{ACKNOWLEDGEMENT}
\label{sec:acknowledgement}
The authors thank the Multi-Robot Systems (MRS) group at Czech Technical University for their support in acquiring valuable real-world forest results.

\printbibliography

\end{document}